\documentclass[11pt,letterpaper]{article}
\usepackage{emnlp2016}
\usepackage{times}
\usepackage{latexsym}
\usepackage{times}
\usepackage{epsfig}
\usepackage{graphicx}
\usepackage{amsmath}
\usepackage{amssymb}
\usepackage{xcolor}
\usepackage{enumerate}
\usepackage{enumitem}
\usepackage[olditem,oldenum]{paralist}
\usepackage{csquotes}
\usepackage{verbatim}
\usepackage[normalem]{ulem}

\usepackage{color}
\usepackage[normalem]{ulem}  
\usepackage{multirow}
\usepackage{rotating}
\usepackage{booktabs}
\usepackage{subcaption}
\usepackage{mysymbols}
\usepackage[breaklinks=true,pagebackref=true,bookmarks=false,colorlinks=true,citecolor=blue]{hyperref}

\emnlpfinalcopy
\DeclareMathOperator*{\argmax}{arg\,max}
\title{Sort Story: Sorting Jumbled Images and Captions into Stories}

\author{Harsh Agrawal\thanks{\,\,\,Denotes equal contribution.}$\,\,^{,1}$ \qquad Arjun Chandrasekaran$^{*,1,}\thanks{\,\,\,Part of this work was done during an internship at TTIC.}$ \\ 
\textbf{Dhruv Batra}$^{3,1}$ \qquad \textbf{Devi Parikh}$^{3,1}$ \qquad \textbf{Mohit Bansal}$^{4,2}$ \vspace{0.005\textwidth} \\
$^1${\fontsize{11}{12}\selectfont Virginia Tech} \hfill $^2${\fontsize{11}{12}\selectfont TTI-Chicago} \hfill 
$^3${\fontsize{11}{12}\selectfont Georgia Institute of Technology} \hfill 
$^4${\fontsize{11}{12}\selectfont UNC Chapel Hill} \\
{\tt\small \{harsh92, carjun, dbatra, parikh\}@vt.edu}, \qquad {\tt\small mbansal@cs.unc.edu}}

\begin{document}

\maketitle
\begin{abstract}
Temporal common sense has applications in AI tasks such as QA, multi-document summarization, and human-AI communication. We propose the task of \emph{sequencing} -- given a jumbled set of aligned image-caption pairs that belong to a story, the task is to sort them such that the output sequence forms a coherent story. We present multiple approaches, via unary (position) and pairwise (order) predictions, and their ensemble-based combinations, achieving strong results on this task. We use both text-based and image-based features, which depict complementary improvements. Using qualitative examples, we demonstrate that our models have learnt interesting aspects of temporal common sense.
\end{abstract}
\vspace{-2pt}
\section{Introduction}
Sequencing is a task for children that is aimed at improving understanding of the temporal occurrence of a sequence of events. The task is, given a jumbled set of images (and maybe captions) that belong to a single story, sort them into the correct order so that they form a coherent story. Our motivation in this work is to enable AI systems to better understand and predict the temporal nature of events in the world. To this end, we train machine learning models to perform the task of ``sequencing''.\par
Temporal reasoning has a number of applications such as multi-document summarization of multiple sources of, say, news information where the relative order of events can be useful to accurately merge information in a temporally consistent manner. In question answering tasks~\cite{richardson2013mctest,fader2014open,weston2015towards,ren2015exploring}, answering questions related to when an event occurs, or what events occurred prior to a particular event require temporal reasoning. A good temporal model of events in everyday life, i.e., a ``temporal common sense'', could also improve the quality of communication between AI systems and humans. \par 

\begin{figure}
    \centering
    \includegraphics[scale=0.083]{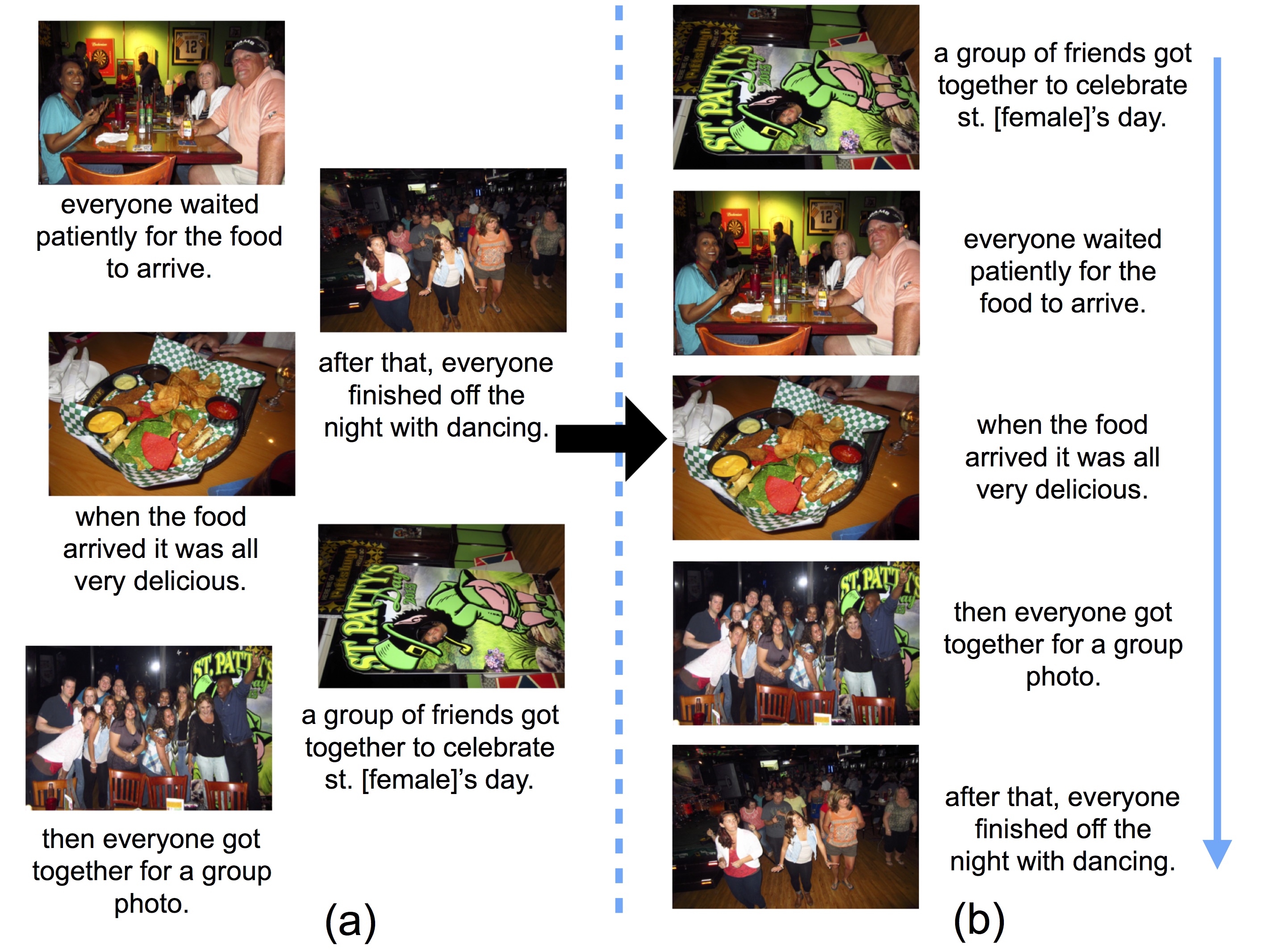}
    \caption{(a) The input is a jumbled set of aligned image-caption pairs. (b) Actual output of the system -- an ordered sequence of image-caption pairs that form a coherent story.}
    \label{fig:teaser}
    \vspace{-9pt}
\end{figure}

Stories are a form of narrative sequences that have an inherent temporal common sense structure. We propose the use of visual stories depicting personal events to learn temporal common sense. We use stories from the Sequential Image Narrative Dataset (SIND)~\cite{visualstorytelling} in which a set of 5 aligned image-caption pairs together form a coherent story. Given an input story that is jumbled (Fig.~\ref{fig:teaser}(a)), we train machine learning models to sort them into a coherent story (Fig.~\ref{fig:teaser}(b)).\footnote{Note that `jumbled' here refers to the loss of temporal ordering; image-caption pairs are still aligned.} \par

Our contributions are as follows: \newline 
-- We propose the task of visual story sequencing. 
\newline 
-- We implement two approaches to solve the task: one based on individual story elements to predict position, and the other based on pairwise story elements to predict relative order of story elements. We also  combine these approaches in a voting scheme that outperforms the individual methods. \newline 
-- As features, we represent a story element as both text-based features from the caption and image-based features, and show that they provide complementary improvements. For text-based features, we use both sentence context and relative order based distributed representations.
\newline 
-- We show qualitative examples of our models learning temporal common sense. 
\section{Related Work}
Temporal ordering has a rich history in NLP research. Scripts \cite{schank2013scripts}, and more recently, narrative chains~\cite{chambers2008unsupervised} contain information about the participants and causal relationships between events that enable the understanding of stories. A number of works~ \cite{mani2005temporally,mani2006machine,boguraev2005timeml} learn temporal relations and properties of news events from the dense, expert-annotated TimeBank corpus \cite{pustejovsky2003timebank}. In our work, however, we use multi-modal story data that has no temporal annotations. \par 
A number of works also reason about temporal ordering by using manually defined linguistic cues \cite{webber1988tense,passonneau1988computational,lapata2006learning,hitzeman1995algorithms,kehler2000coherence}. Our approach uses neural networks to avoid feature design for learning temporal ordering. \par 

Recent works~\cite{modiCONLL2014,modiCONLL2016} learn distributed representations for predicates in a sentence for the tasks of event ordering and cloze evaluation. Unlike their work, our approach makes use of multi-modal data with \emph{free-form} natural language text to learn event embeddings. Further, our models are trained end-to-end while their pipelined approach involves parsing and extracting verb frames from each sentence, where errors may propagate from one module to the next (as discussed in \secref{sec:results}). \par 

\newcite{chen2009content} use a generalized Mallows model for modeling sequences for coherence within single documents. Their approach may also be applicable to our task.
Recently,~\newcite{mostafazadeh2016corpus} presented the ``ROCStories'' dataset of 5-sentence stories with stereotypical causal and temporal relations between events. In our work though, we make use of a multi-modal story-dataset that contains \emph{both} images and associated story-like captions. \par
Some works in vision~\cite{pickup2014seeing,basha2012photo} also temporally order images; typically by finding correspondences between multiple images of the same scene using geometry-based approaches. Similarly, \newcite{choi2016video} compose a story out of multiple short video clips. They define metrics based on scene dynamics and coherence, and use dense optical flow and patch-matching. In contrast, our work deals with stories containing potentially visually dissimilar but \emph{semantically} coherent set of images and captions. \par

A few other recent works~\cite{kim2015joint,kim2014joint,kim2014reconstructing,sigurdsson2016learning,bosselut-event-script,wanglow} summarize hundreds of individual streams of information (images, text, videos) from the web that deal with a single concept or event, to learn a common theme or \emph{storyline} or for \emph{timeline summarization}. 
Our task, however, is to predict the correct sorting of a given story, which is different from summarization or retrieval.
\newcite{ramanathan2015learning} attempt to learn temporal embeddings of video frames in complex events. While their motivation is similar to ours, they deal with sampled frames from a video while we attempt to learn temporal common sense from \emph{multi-modal} stories consisting of a sequence of aligned image-caption pairs.
\vspace{-0.25pt}
\section{Approach}
\label{sec:approach}
In this section, we first describe the two components in our approach: unary scores that do not use context, and pairwise scores that encode relative orderings of elements. Next, we describe how we combine these scores through a voting scheme.

\subsection{Unary Models}
Let $\sigma \in \Sigma_n$ denote a permutation 
of $n$ elements (image-caption pairs). We use $\sigma_i$ 
to denote the position of element $i$ in the permutation $\sigma$. 
A unary score $S_u(\sigma)$ captures the appropriateness of each story element $i$ in position $\sigma_i$:
\begin{equation}
	S_u(\sigma) =  \sum\limits_{i=1}^{n}P(\sigma_i| i)
\end{equation}
where $P(\sigma_i|i)$ denotes the probability of the element $i$ being present in position $\sigma_i$,  
which is the output from an $n$-way softmax layer 
in a deep neural network. 
We experiment with 2 networks -- \\ 
\noindent(1) A language-alone unary model (Skip-Thought+MLP) that uses a Gated Recurrent Unit (GRU) proposed by~\newcite{cho2014learning} to 
embed a caption into a vector space. We use the 
Skip-Thought~\cite{kiros2015skip} GRU, which is trained on the BookCorpus~\cite{zhu2015aligning} to predict the context (preceding and following sentences) of a given sentence. These embeddings are fed as input into a Multi-Layer Perceptron (MLP). \\
\noindent(2) A language+vision unary model (Skip-Thought+CNN+MLP) that 
embeds the caption as above and embeds the image via a Convolutional Neural Network (CNN). We use the activations from the penultimate layer of the 19-layer VGG-net~\cite{simonyan2014very}, which have been shown to generalize well. Both embeddings are concatenated and fed as input to an MLP. \par
In both cases, the best ordering of the story elements (optimal 
permutation) $\sigma^{*} = \argmax_{\sigma \in \Sigma_n} S_u(\sigma)$ can be found efficiently in $O(n^3)$ time with the Hungarian algorithm~\cite{munkres1957algorithms}. Since these unary scores are not influenced by other elements in the story, they capture the semantics and linguistic structures associated with  specific positions of stories \emph{e.g.}, the beginning, the middle, and the end. \par 

\subsection{Pairwise Models}
Similar to learning to rank approaches~\cite{hang2011short}, we develop pairwise scoring models that given a pair of elements $(i,j)$, learn to assign a score: \\ $S(\ind{\sigma_i < \sigma_j} \mid i,j)$ indicating whether element $i$ should be placed before element $j$ in the permutation $\sigma$. 
Here, $\ind{\cdot}$ indicates the Iverson bracket (which is $1$ if the input argument is true and 0 otherwise). We develop and experiment with the following 3 pairwise models: \\
\noindent(1) A language-alone pairwise model (Skip-Thought+MLP) that takes as input a pair of Skip-Thought embeddings and trains an MLP (with hinge-loss) that outputs $S(\ind{\sigma_i < \sigma_j} \mid i,j)$, 
the score for placing $i$ before $j$. \\
\noindent(2) A language+vision pairwise model (Skip-Thought+CNN+MLP) 
that concatenates the Skip-Thought and CNN embeddings for $i$ 
and $j$ and trains a similar MLP as above. \\
\noindent(3) A language-alone neural position embedding (NPE) model. Instead of using frozen Skip-Thought embeddings, we learn a task-aware ordered distributed embedding for sentences. Specifically, each sentence in the story is embedded $X = (\xb_1, \ldots, \xb_n), \,\,  \xb_i \in \RF^d_+$, via an LSTM \cite{hochreiter1997long} with ReLU non-linearities. Similar to the max-margin loss that is applied to negative examples by~\newcite{vendrov2015order}, we use an asymmetric penalty that encourages sentences appearing early in the story to be placed closer to the origin than sentences appearing later in the story. 
\newcommand\norm[1]{\Big\lVert#1\Big\rVert}
\begin{equation}
\label{eq:npe}
  \begin{aligned}
    L_{ij} &= \norm{\max(0, \alpha - (\xb_j-\xb_i))}^{2}\\
    Loss &= \sum\limits_{1<=i<j=n}L_{ij}
  \end{aligned}
  \end{equation}
At train time, the parameters of the LSTM are learned end-to-end to minimize this asymmetric ordered loss (as measured over the gold-standard sequences). At test time, we use $S(\ind{\sigma_i < \sigma_j} \mid i,j) = L_{ij}$. Thus, as we move away from the origin in the embedding space, we  traverse through the sentences in a story. Each of these three pairwise approaches assigns a score $S(\sigma_i,\sigma_j | i,j)$ to an ordered pair of elements (i,j), which is used to construct a pairwise scoring model: 
{\small
\begin{equation}
S_p(\sigma) = \hspace{-15pt} \sum_{1<=i<j<=n} \Big\{ 
S(\ind{\sigma_i < \sigma_j}) - S(\ind{\sigma_j < \sigma_i}) \Big\},
\label{eq:pairwise}
\end{equation}
}
by summing over the scores for all possible ordered pairs 
in the permutation. 
This pairwise score captures local contextual information in stories. Finding the best permutation 
$\sigma^{*} = \argmax_{\sigma \in \Sigma_n} S_p(\sigma)$
under this pairwise model is NP-hard so approximations will be required. In our 
experiments, we study short sequences ($n=5$), where 
the space of permutations is easily enumerable 
($5!=120$). For longer sequences, we can 
utilize integer programming methods or well-studied 
spectral relaxations for this problem. 

\subsection{Voting-based Ensemble}
\label{exhaustive}
To combine the complementary information captured by the unary  ($S_u$) and pairwise models ($S_p$), we use a voting-based ensemble. For each method in the ensemble, we find the top three permutations. Each of these permutations $(\sigma^k)$ then vote for a particular element to be placed at a particular position. Let $V$ be a vote matrix such that $V_{ij}$ stores the number of votes for $i^{th}$ element to occur at $j^{th}$ position, \ie 
$ V_{ij} = \sum_{k} \ind{\sigma^k_i == j})$. 
We use the Hungarian algorithm to find the optimal permutation that maximizes the votes assigned, 
\ie $\sigma^*_{\text{vote}} = \argmax_{\sigma \in \Sigma_n} 
\sum_{i=1}^n \sum_{j=1}^n V_{ij} \cdot \ind{\sigma_i == j}$. We experimented with a number of model voting combinations and found the combination of pairwise Skip-Thought+CNN+MLP and neural position embeddings to work best (based on a validation set). 
\vspace*{5mm}
\section{Experiments}
\label{sec:exp}
\subsection{Data}
\label{ssec:data} 
We train and evaluate our model on personal multi-modal stories from the SIND (Sequential Image Narrative Dataset)~\cite{visualstorytelling}, where each story is a sequence of 5 images and corresponding story-like captions. The narrative captions in this dataset, e.g., ``friends having a good time'' (as opposed to ``people sitting next to each other'') capture a sequential, conversational language, which is characteristic of stories. We use 40,155 stories for training, 4990 for validation and 5055 stories for testing.
\subsection{Metrics}
\label{sec:metrics}
We evaluate the performance of our model at correctly ordering a jumbled set of story elements using the following 3 metrics: 
\newline 
\textbf{Spearman's rank correlation} (Sp.) \cite{spearman1904proof} measures if the ranking of story elements in the predicted and ground truth orders are monotonically related (higher is better).  \newline
\textbf{Pairwise accuracy} (Pairw.) measures the fraction of pairs of elements whose predicted relative ordering is the same as the ground truth order (higher is better). 
\newline 
\textbf{Average Distance} (Dist.) measures the average change in position of all elements in the predicted story from their respective positions in the ground truth story (lower is better).

\subsection{Results}
\label{sec:results}

\begin{table}[t]
\setlength{\tabcolsep}{2pt}
{\small
\begin{center}
\resizebox{\columnwidth}{!}{
\begin{tabular}{@{}llcccc@{}}
\toprule
\textbf{Method} & \textbf{Features} & \textbf{Sp.} & \textbf{Pairw.} & \textbf{Dist.} \\ 
\midrule
Random Order &  & \multicolumn{1}{r}{0.000} & \multicolumn{1}{r}{0.500} & \multicolumn{1}{r}{1.601} \\
\midrule
Unary& SkipThought & \multicolumn{1}{r}{0.508} & \multicolumn{1}{r}{0.718} & \multicolumn{1}{r}{1.373} \\
& SkipThought + Image  & \multicolumn{1}{r}{0.532} & \multicolumn{1}{r}{0.729} & \multicolumn{1}{r}{1.352}\\
\midrule
 Pairwise& SkipThought & \multicolumn{1}{r}{0.546} & \multicolumn{1}{r}{0.732} & \multicolumn{1}{r}{0.923} \\
& SkipThought + Image & \multicolumn{1}{r}{0.565} & \multicolumn{1}{r}{0.740} & \multicolumn{1}{r}{0.897} \\
\midrule
 Pairwise Order \ \ & NPE & \multicolumn{1}{r}{0.480} & \multicolumn{1}{r}{0.704} & \multicolumn{1}{r}{1.010} \\
\midrule
 Voting & SkipThought + Image \ \ & \multicolumn{1}{r}{\textbf{0.675}} & \multicolumn{1}{r}{\textbf{0.799}} & \multicolumn{1}{r}{\textbf{0.724}} \\
 &(Pairwise) + NPE &  & &  \\
\bottomrule
\end{tabular}}
\end{center}
\caption{Performance of our different models and features at the sequencing task.}
\label{tab:peformance}
}
\end{table}
\paragraph{Pairwise Models vs Unary Models} 
As shown in Table~\ref{tab:peformance}, the pairwise models based on Skip-Thought features outperform the unary models in our task. However, the Pairwise Order Model performs worse than the unary Skip-Thought model, suggesting that the Skip-Thought features, which encode context of a sentence, also provide a crucial signal for temporal ordering of story sentences. 
\paragraph{Contribution of Image Features} 
Augmenting the text features with image features results in a visible performance improvement of both the model trained with unary features and the model trained with pairwise features.
While image features by themselves result in poor performance on this task, they seem to capture temporal information that is complementary to the text features. 

\begin{figure*}[t!]
    \centering
    \begin{subfigure}[b]{0.25\textwidth}
        \centering
        \includegraphics[width=\textwidth]{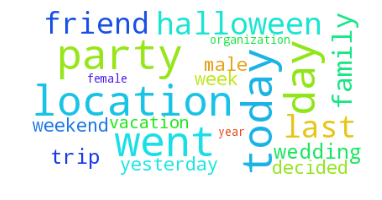}
        \vspace*{-7mm}
        \caption{First Position \label{subfig:wc_1}}
    \end{subfigure}
    \begin{subfigure}[b]{0.25\textwidth}
        \centering
        \includegraphics[width=\textwidth]{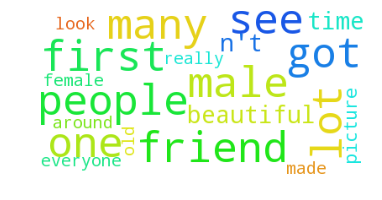}
		\vspace*{-7mm}
        \caption{Second Position \label{subfig:wc_2}}
    \end{subfigure}
    \begin{subfigure}[b]{0.25\textwidth}
        \centering
        \includegraphics[width=\textwidth]{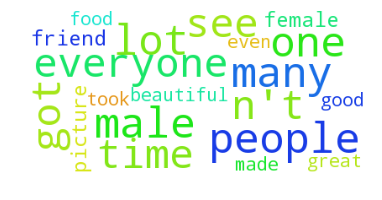}        
        \vspace*{-7mm}
        \caption{Third Position \label{subfig:wc_3}}
    \end{subfigure}
    \\
    \begin{subfigure}[b]{0.25\textwidth}
        \centering
        \includegraphics[width=\textwidth]{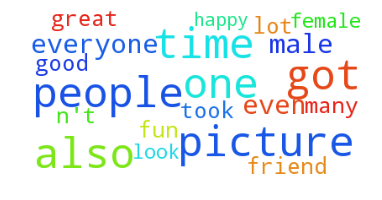}        
        \vspace*{-7mm}
        \caption{Fourth Position \label{subfig:wc_4}}
    \end{subfigure}
    \begin{subfigure}[b]{0.25\textwidth}
        \centering
        \includegraphics[width=\textwidth]{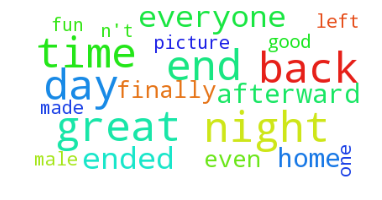}        
        \vspace*{-7mm}
        \caption{Fifth Position \label{subfig:wc_5}}
    \end{subfigure}
    \caption{Word cloud corresponding to most discriminative words for each position.}
    \label{fig:word_cloud}
\end{figure*}

\paragraph{Ensemble Voting}
To exploit the fact that unary and pairwise models, as well as text and image features, capture different aspects of the story, we combine them using a voting ensemble. Based on the validation set, we found that combining the Pairwise Order model and the Pairwise model with both Skip-Thought and Image (CNN) features performs the best. This voting based method achieves the best performance on all three metrics. This shows that our different approaches indeed capture complementary information regarding feasible orderings of caption-image pairs to form a coherent story. \par 

For comparison to existing related work, we tried to duplicate the pipelined approach of \newcite{modiCONLL2014}. For this, we first parse our story sentences to extract SVO (subject, verb, object) tuples (using the Stanford Parser~\cite{chen2014fast}). However, this step succeeds for only 60\% of our test data. Now even if we consider a \emph{perfect} downstream algorithm that \emph{always} makes the correct position prediction given SVO tuples, the overall performance is still a Spearman correlation of just 0.473, i.e., the \emph{upper bound} performance of this pipelined approach is \emph{lower} than the performance of our text-only end-to-end model (correlation of 0.546) in Table~\ref{tab:peformance}.

\subsection{Qualitative Analysis}
Visualizations of position predictions from our model demonstrate that it has learnt the \emph{three act structure}~\cite{trottier1998screenwriter} in stories -- the setup, the middle and the climax. We also present success and failure examples of our sorting model's predictions. See the supplementary for more details and figures. \par

We visualize our model's \emph{temporal common sense}, in Fig.~\ref{fig:word_cloud}. The word clouds show \emph{discriminative words} -- the words that the model believes are indicative of sentence positions in a story. The size of a word is proportional to the ratio of its frequency of occurring in that position to other positions. Some words like `party', `wedding', etc., probably because our model believes that the start the story describes the setup -- the occasion or event. People often tend to describe meeting friends or family members which probably results in the discriminative words such as `people', `friend', `everyone' in the second and the third sentences. Moreover, the model believes that people tend to conclude the stories using words like `finally', `afterwards', tend to talk about `great day', group `pictures' with everyone, etc. 
\section{Conclusion}
We propose the task of ``sequencing'' in a set of image-caption pairs, with the motivation of learning temporal common sense. We implement multiple neural network models based on individual and pairwise element-based predictions (and their ensemble), and utilize both image and text features, to achieve strong performance on the task. Our best system, on average, predicts the ordering of sentences to within a distance error of 0.8 (out of 5) positions. We also analyze our predictions and show qualitative examples that demonstrate temporal common sense. 
\noindent 
\section*{Acknowledgements} 
We thank Ramakrishna Vedantam and the anonymous reviewers for their helpful suggestions. This work was supported by: NSF CAREER awards to DB and DP, ARO YIP awards to DB and DP, ICTAS Junior Faculty awards to DB and DP, Google Faculty Research award to DP and DB, ARL grant W911NF-15-2-0080 to DP and DB, ONR grant N00014-14-1-0679 to DB and N00014-16-1-2713 to DP, ONR YIP award to DP, Paul G. Allen Family Foundation Allen Distinguished Investigator award to DP, Alfred P. Sloan Fellowship to DP, AWS in Education Research grant to DB, NVIDIA GPU donations to DB and MB, an IBM Faculty Award and Bloomberg Data Science Research Grant to MB.
\appendix
\vspace{15pt}
\noindent
{\fontsize{12}{12}\selectfont \textbf{Appendix}} \par
\section{Confusion Matrix for Predicting Position of an Element}
\label{subsec:confusion}
\begin{figure}[t]
  \centering
  \includegraphics[width=0.4\textwidth]{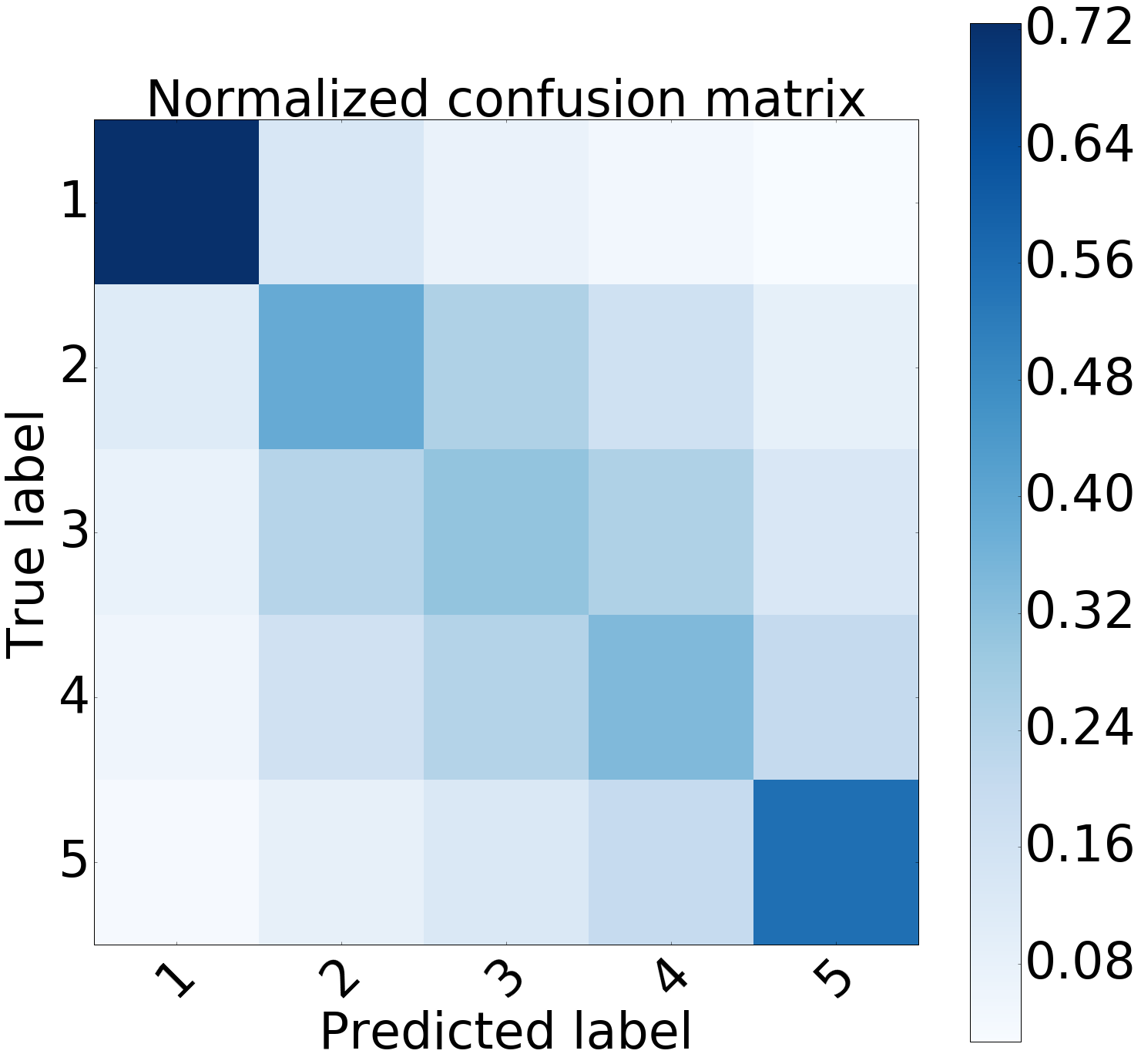}
  \caption{Confusion matrix for predictions from the best performing model i.e Voting ensemble of Pairwise Skip-Thought+image(CNN) and Pairwise Order Neural Position Embedding (NPE).}
  \label{fig:conf_mat}
\end{figure}

We visualize the 5-way classification confusion matrix for our best performing method i.e., Voting ensemble of Pairwise Skip-Thought+Image(CNN) and Pairwise Order (Neural Position Embedding (NPE)) in Fig.~\ref{fig:conf_mat}. 
The block-diagonal matrix structure shows that the model predicts the first and the last element of a story reasonably well but is often confused by elements in the middle of the story. This visualization suggests that the model has learnt the \emph{three act structure} in stories, i.e., the setup, the middle and the climax.

\section{Predicted Stories}
\begin{figure*}
  \centering
  \begin{subfigure}[b]{0.90\textwidth}
    \includegraphics[width=\textwidth]{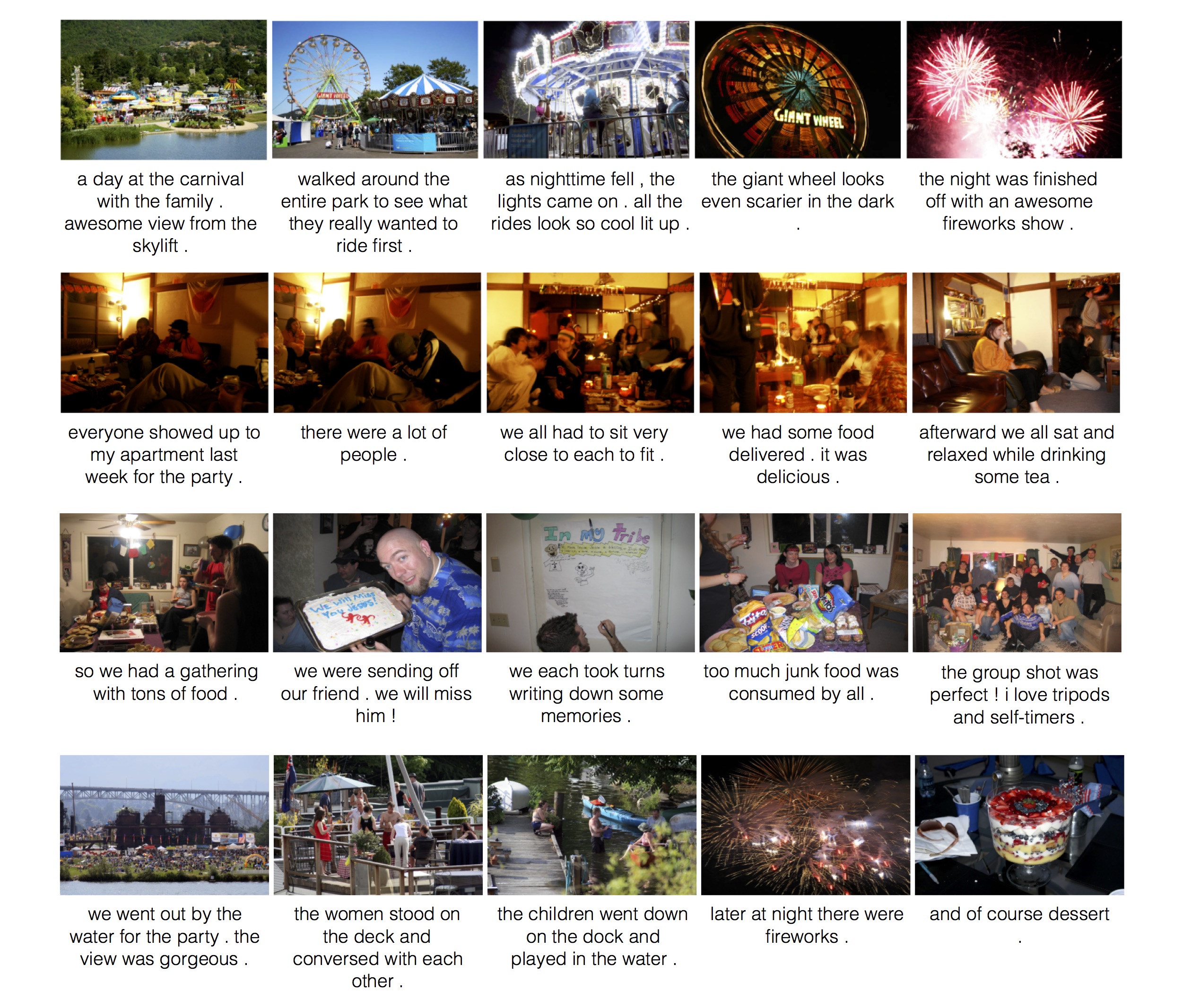}
    \vspace*{-8mm}
    \caption{Examples of stories for which the temporal sequence of elements was predicted perfectly.  \label{fig:pos_ex}}
  \vspace{10pt}
  \end{subfigure}

  \begin{subfigure}[b]{0.90\textwidth}
    \centering
    \includegraphics[width=\textwidth]{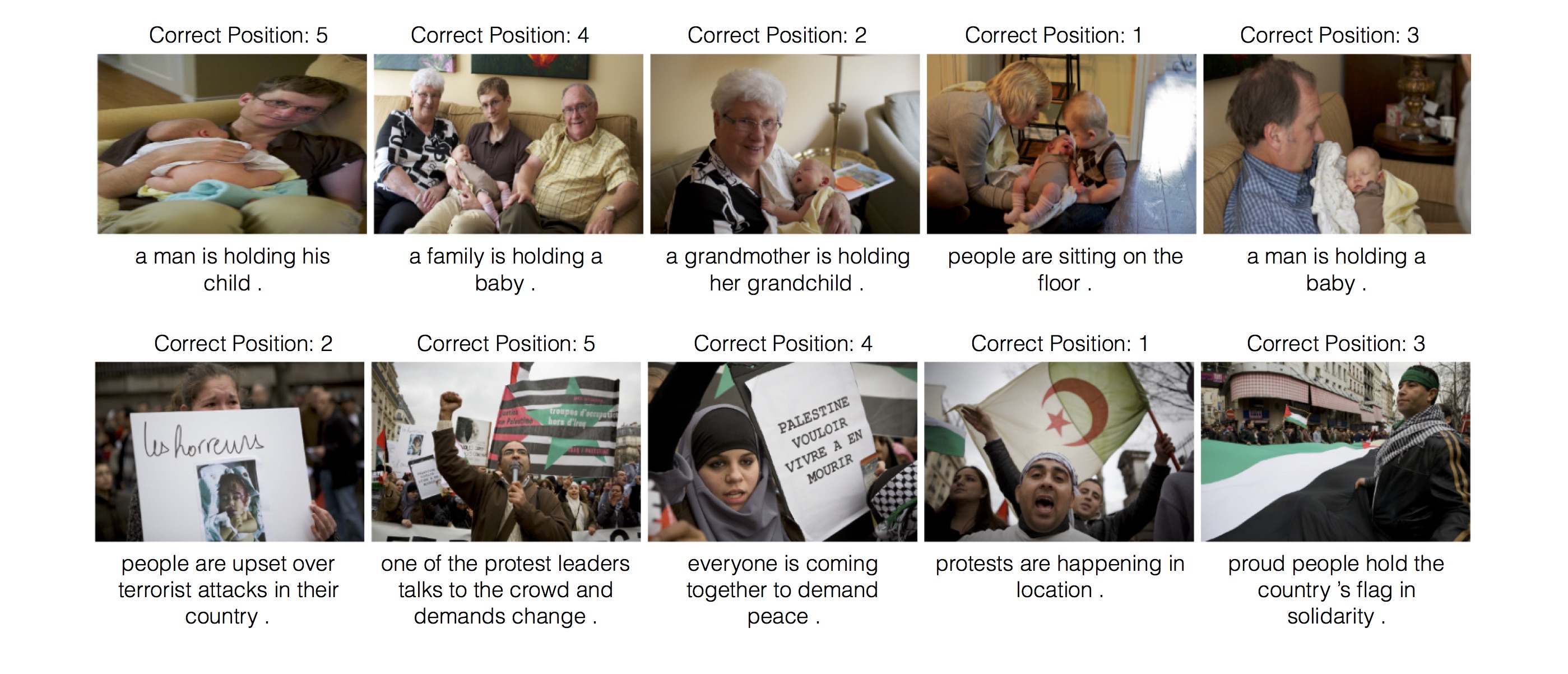}  
    \vspace*{-8mm}
	\caption{Examples of stories for which the model failed to predict the correct position of any story element. The elements (images and captions) in a story are generic, with no clear temporal ordering. The stories seem to lack a coherent narrative.}
    \label{fig:neg_ex}
  \end{subfigure}
  \vspace{5pt}
  \caption{Examples of success and failure cases of temporal order prediction of story elements by our best performing model.\label{fig:qual_ex}}
\end{figure*}

We present qualitative examples of story orders predicted by the best performing model in Fig.~\ref{fig:qual_ex}. Fig.~\ref{fig:pos_ex} shows example stories in which the position of all elements are predicted correctly. Fig.~\ref{fig:neg_ex} shows stories in which none of the positions are predicted correctly by our model. These two examples show that our model clearly fails when there is no inherent temporal order in the story either via language or images.

\section{Temporal Common Sense}
\label{subsec:commonsense}
\begin{figure*}[t]
    \centering
    \vspace{0pt}
    \begin{subfigure}[b]{0.30\textwidth}
        \centering
        \includegraphics[width=\textwidth]{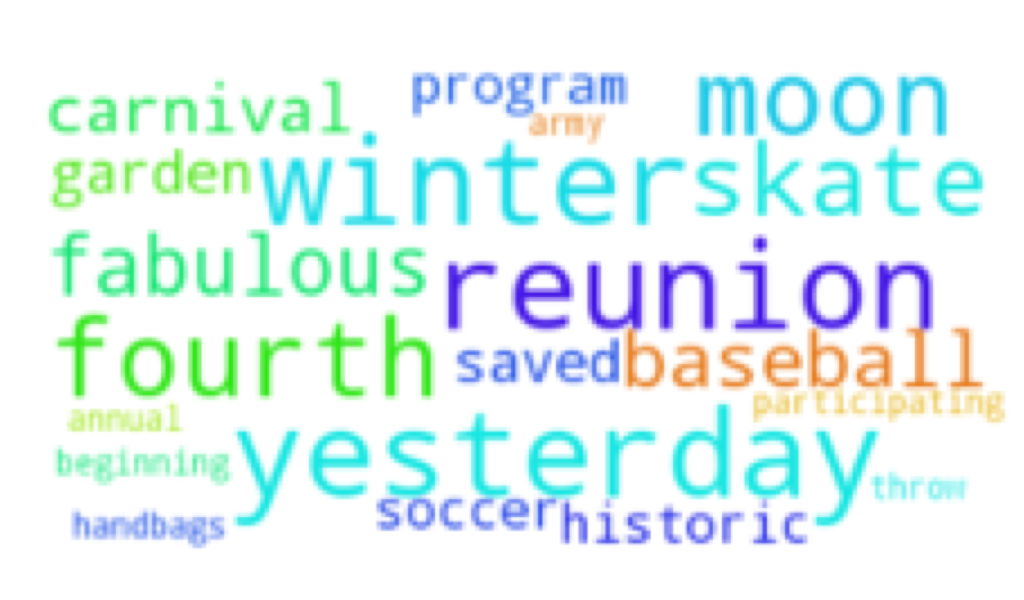}
        \caption{First Position \label{subfig:w_1}}
    \end{subfigure}
    \begin{subfigure}[b]{0.30\textwidth}
        \centering
        \includegraphics[width=\textwidth]{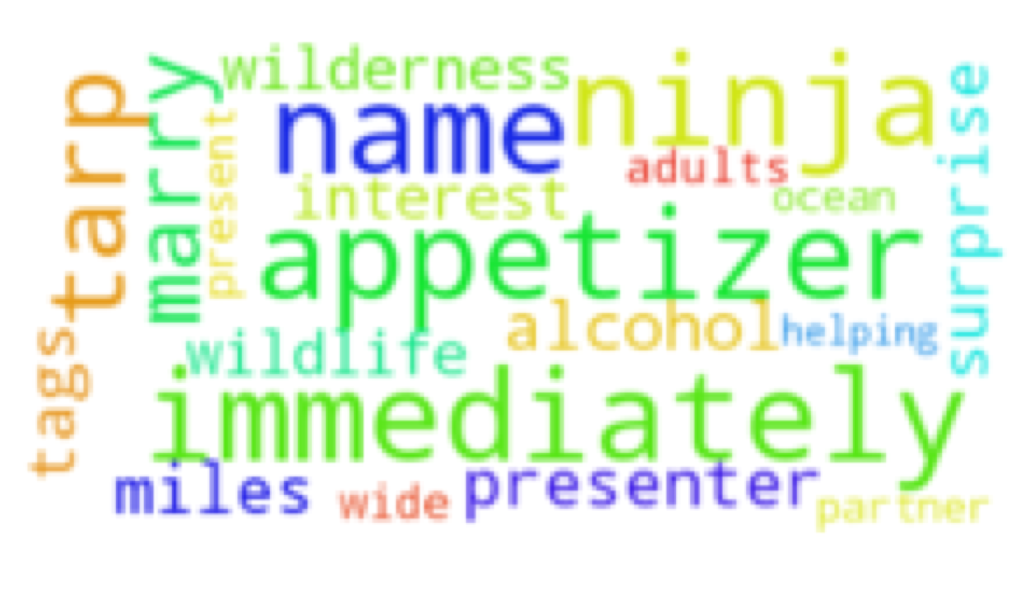}
        \caption{Second Position \label{subfig:w_2}}
    \end{subfigure}
    \begin{subfigure}[b]{0.30\textwidth}
        \centering
        \includegraphics[width=\textwidth]{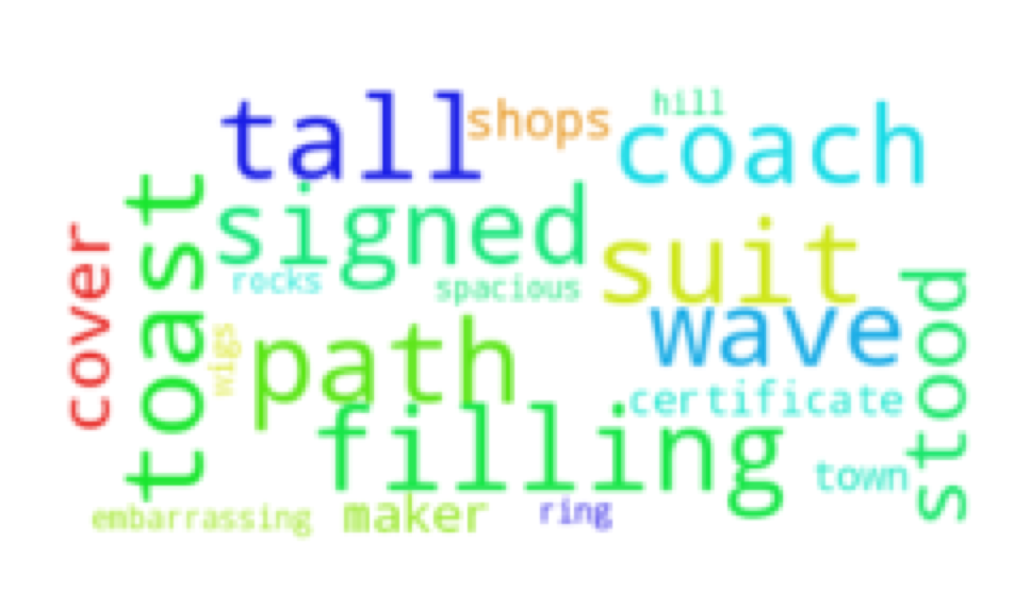}
        \caption{Third Position \label{subfig:w_3}}
    \end{subfigure}
    \\
    \begin{subfigure}[b]{0.30\textwidth}
        \centering
        \includegraphics[width=\textwidth]{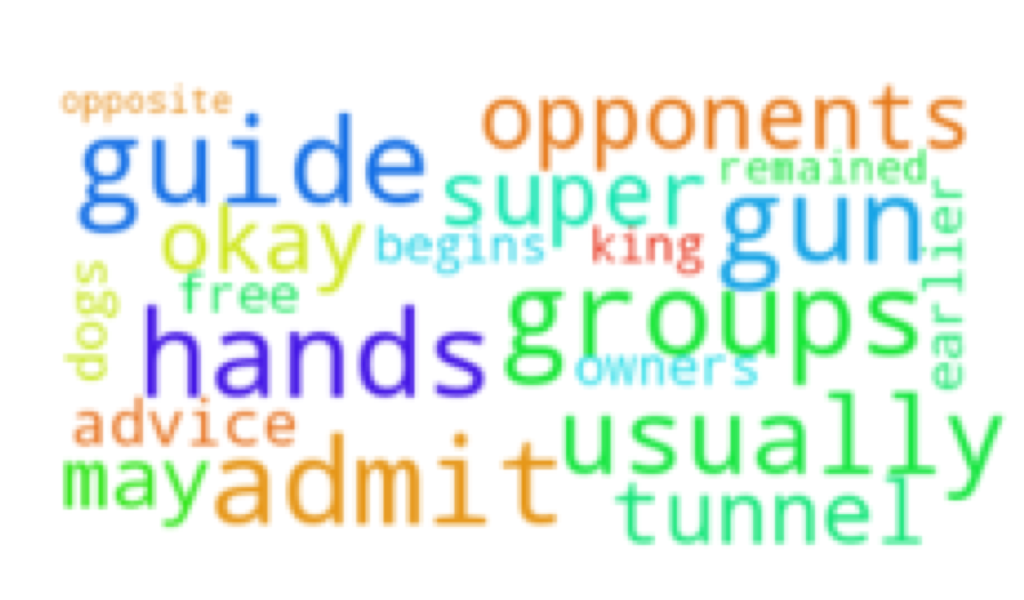}
        \caption{Fourth Position \label{subfig:wc_4}}
    \end{subfigure}
    \begin{subfigure}[b]{0.30\textwidth}
        \centering
        \includegraphics[width=\textwidth]{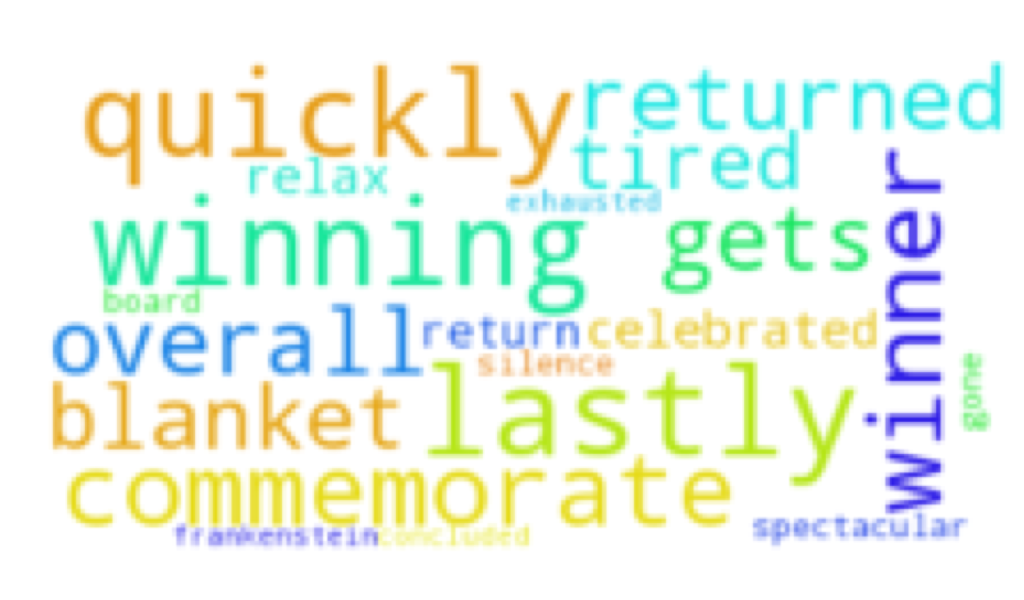}
        \caption{Fifth Position \label{subfig:wc_5}}
    \end{subfigure}
    \caption{Discriminative words in each position of all correctly predicted stories.}
    \label{fig:dic_pos_neg_word_cloud}
\end{figure*}

In the word cloud in Fig.~\ref{fig:dic_pos_neg_word_cloud}, we visualize the words that the model finds \emph{discriminative} in correct predictions. These are words from \emph{correctly} predicted stories that the model believes are indicative of sentence positions in a story. The size of a word is proportional to the ratio of its frequency of occurring in that position to other positions. Our model captures events such as `carnival', `reunion', and sports topics like `baseball', `soccer', `skate' in the first position. This could be the case because the first sentence of a story usually introduces the event that the story is based on. In Fig.~\ref{subfig:wc_5} (word-cloud of the last sentence), we also observe that the model correctly learns cue-words such as `overall', and `lastly'. It also learns words and events that frequently conclude stories such as `returned', `tired', `winning', `winner', and `celebration'.

\bibliographystyle{emnlp2016}
\bibliography{emnlp2016}

\begin{thebibliography}{}

\bibitem[\protect\citename{Basha \bgroup et al.\egroup }2012]{basha2012photo}
Tali Basha, Yael Moses, and Shai Avidan.
\newblock 2012.
\newblock Photo sequencing.
\newblock In {\em ECCV}.

\bibitem[\protect\citename{Boguraev and Ando}2005]{boguraev2005timeml}
Branimir Boguraev and Rie~Kubota Ando.
\newblock 2005.
\newblock Timeml-compliant text analysis for temporal reasoning.
\newblock In {\em IJCAI}.

\bibitem[\protect\citename{Bosselut \bgroup et al.\egroup
  }2016]{bosselut-event-script}
Antoine Bosselut, Jianfu Chen, David Warren, Hannaneh Hajishirzi, and Yejin
  Choi.
\newblock 2016.
\newblock Learning prototypical event structure from photo albums.
\newblock In {\em ACL}.

\bibitem[\protect\citename{Chambers and
  Jurafsky}2008]{chambers2008unsupervised}
Nathanael Chambers and Daniel Jurafsky.
\newblock 2008.
\newblock Unsupervised learning of narrative event chains.
\newblock In {\em ACL}. Citeseer.

\bibitem[\protect\citename{Chen and Manning}2014]{chen2014fast}
Danqi Chen and Christopher~D Manning.
\newblock 2014.
\newblock A fast and accurate dependency parser using neural networks.
\newblock In {\em EMNLP}.

\bibitem[\protect\citename{Chen \bgroup et al.\egroup }2009]{chen2009content}
Harr Chen, SRK Branavan, Regina Barzilay, David~R Karger, et~al.
\newblock 2009.
\newblock Content modeling using latent permutations.
\newblock {\em Journal of Artificial Intelligence Research}.

\bibitem[\protect\citename{Cho \bgroup et al.\egroup }2014]{cho2014learning}
Kyunghyun Cho, Bart Van~Merri{\"e}nboer, Caglar Gulcehre, Dzmitry Bahdanau,
  Fethi Bougares, Holger Schwenk, and Yoshua Bengio.
\newblock 2014.
\newblock Learning phrase representations using rnn encoder-decoder for
  statistical machine translation.
\newblock In {\em EMNLP}.

\bibitem[\protect\citename{Choi \bgroup et al.\egroup }2016]{choi2016video}
Jinsoo Choi, Tae-Hyun Oh, and In~So~Kweon.
\newblock 2016.
\newblock Video-story composition via plot analysis.
\newblock In {\em CVPR}.

\bibitem[\protect\citename{Fader \bgroup et al.\egroup }2014]{fader2014open}
Anthony Fader, Luke Zettlemoyer, and Oren Etzioni.
\newblock 2014.
\newblock Open question answering over curated and extracted knowledge bases.
\newblock In {\em ACM SIGKDD}.

\bibitem[\protect\citename{Hang}2011]{hang2011short}
LI~Hang.
\newblock 2011.
\newblock A short introduction to learning to rank.
\newblock {\em IEICE TRANSACTIONS on Information and Systems}.

\bibitem[\protect\citename{Hitzeman \bgroup et al.\egroup
  }1995]{hitzeman1995algorithms}
Janet Hitzeman, Marc Moens, and Claire Grover.
\newblock 1995.
\newblock Algorithms for analysing the temporal structure of discourse.
\newblock In {\em EACL}.

\bibitem[\protect\citename{Hochreiter and Schmidhuber}1997]{hochreiter1997long}
Sepp Hochreiter and J{\"u}rgen Schmidhuber.
\newblock 1997.
\newblock Long short-term memory.
\newblock {\em Neural computation}.

\bibitem[\protect\citename{Huang \bgroup et al.\egroup
  }2016]{visualstorytelling}
Ting-Hao~(Kenneth) Huang, Francis Ferraro, Nasrin Mostafazadeh, Ishan Mishra,
  Aishwarya Agrawal, Jacob Devlin, Ross Girshick, Xiaodong He, Pushmeet Kohli,
  Dhruv Batra, C.~Lawrence Zitnick, Devi Parikh, Lucy Vanderwende, Michel
  Galley, and Margaret Mitchell.
\newblock 2016.
\newblock Visual storytelling.
\newblock In {\em NAACL}.

\bibitem[\protect\citename{Kehler}2000]{kehler2000coherence}
Andrew Kehler.
\newblock 2000.
\newblock Coherence and the resolution of ellipsis.
\newblock {\em Linguistics and Philosophy}.

\bibitem[\protect\citename{Kim and Xing}2014]{kim2014reconstructing}
Gunhee Kim and Eric Xing.
\newblock 2014.
\newblock Reconstructing storyline graphs for image recommendation from web
  community photos.
\newblock In {\em CVPR}.

\bibitem[\protect\citename{Kim \bgroup et al.\egroup }2014]{kim2014joint}
Gunhee Kim, Leonid Sigal, and Eric Xing.
\newblock 2014.
\newblock Joint summarization of large-scale collections of web images and
  videos for storyline reconstruction.
\newblock In {\em CVPR}.

\bibitem[\protect\citename{Kim \bgroup et al.\egroup }2015]{kim2015joint}
Gunhee Kim, Seungwhan Moon, and Leonid Sigal.
\newblock 2015.
\newblock Joint photo stream and blog post summarization and exploration.
\newblock In {\em CVPR}.

\bibitem[\protect\citename{Kiros \bgroup et al.\egroup }2015]{kiros2015skip}
Ryan Kiros, Yukun Zhu, Ruslan~R Salakhutdinov, Richard Zemel, Raquel Urtasun,
  Antonio Torralba, and Sanja Fidler.
\newblock 2015.
\newblock Skip-thought vectors.
\newblock In {\em NIPS}.

\bibitem[\protect\citename{Lapata and Lascarides}2006]{lapata2006learning}
Mirella Lapata and Alex Lascarides.
\newblock 2006.
\newblock Learning sentence-internal temporal relations.
\newblock {\em Journal of Artificial Intelligence Research}.

\bibitem[\protect\citename{Mani and Schiffman}2005]{mani2005temporally}
Inderjeet Mani and Barry Schiffman.
\newblock 2005.
\newblock Temporally anchoring and ordering events in news.
\newblock {\em Time and Event Recognition in Natural Language. John Benjamins}.

\bibitem[\protect\citename{Mani \bgroup et al.\egroup }2006]{mani2006machine}
Inderjeet Mani, Marc Verhagen, Ben Wellner, Chong~Min Lee, and James
  Pustejovsky.
\newblock 2006.
\newblock Machine learning of temporal relations.
\newblock In {\em COLING-ACL}.

\bibitem[\protect\citename{Modi and Titov}2014]{modiCONLL2014}
Ashutosh Modi and Ivan Titov.
\newblock 2014.
\newblock Inducing neural models of script knowledge.
\newblock In {\em CoNLL}.

\bibitem[\protect\citename{Modi}2016]{modiCONLL2016}
Ashutosh Modi.
\newblock 2016.
\newblock Event embeddings for semantic script modeling.
\newblock In {\em CoNLL}.

\bibitem[\protect\citename{Mostafazadeh \bgroup et al.\egroup
  }2016]{mostafazadeh2016corpus}
Nasrin Mostafazadeh, Nathanael Chambers, Xiaodong He, Devi Parikh, Dhruv Batra,
  Lucy Vanderwende, Pushmeet Kohli, and James Allen.
\newblock 2016.
\newblock A corpus and cloze evaluation for deeper understanding of commonsense
  stories.
\newblock In {\em NAACL}.

\bibitem[\protect\citename{Munkres}1957]{munkres1957algorithms}
James Munkres.
\newblock 1957.
\newblock Algorithms for the assignment and transportation problems.
\newblock {\em Journal of the Society for Industrial and Applied Mathematics}.

\bibitem[\protect\citename{Passonneau}1988]{passonneau1988computational}
Rebecca~J Passonneau.
\newblock 1988.
\newblock A computational model of the semantics of tense and aspect.
\newblock {\em Computational Linguistics}.

\bibitem[\protect\citename{Pickup \bgroup et al.\egroup
  }2014]{pickup2014seeing}
Lyndsey Pickup, Zheng Pan, Donglai Wei, YiChang Shih, Changshui Zhang, Andrew
  Zisserman, Bernhard Scholkopf, and William Freeman.
\newblock 2014.
\newblock Seeing the arrow of time.
\newblock In {\em CVPR}.

\bibitem[\protect\citename{Pustejovsky \bgroup et al.\egroup
  }2003]{pustejovsky2003timebank}
James Pustejovsky, Patrick Hanks, Roser Sauri, Andrew See, Robert Gaizauskas,
  Andrea Setzer, Dragomir Radev, Beth Sundheim, David Day, Lisa Ferro, et~al.
\newblock 2003.
\newblock The timebank corpus.
\newblock In {\em Corpus linguistics}.

\bibitem[\protect\citename{Ramanathan \bgroup et al.\egroup
  }2015]{ramanathan2015learning}
Vignesh Ramanathan, Kevin Tang, Greg Mori, and Li~Fei-Fei.
\newblock 2015.
\newblock Learning temporal embeddings for complex video analysis.
\newblock In {\em CVPR}.

\bibitem[\protect\citename{Ren \bgroup et al.\egroup }2015]{ren2015exploring}
Mengye Ren, Ryan Kiros, and Richard Zemel.
\newblock 2015.
\newblock Exploring models and data for image question answering.
\newblock In {\em NIPS}.

\bibitem[\protect\citename{Richardson \bgroup et al.\egroup
  }2013]{richardson2013mctest}
Matthew Richardson, Christopher~JC Burges, and Erin Renshaw.
\newblock 2013.
\newblock Mctest: A challenge dataset for the open-domain machine comprehension
  of text.
\newblock In {\em EMNLP}.

\bibitem[\protect\citename{Schank and Abelson}2013]{schank2013scripts}
Roger~C Schank and Robert~P Abelson.
\newblock 2013.
\newblock {\em Scripts, plans, goals, and understanding: An inquiry into human
  knowledge structures}.
\newblock Psychology Press.

\bibitem[\protect\citename{Sigurdsson \bgroup et al.\egroup
  }2016]{sigurdsson2016learning}
Gunnar~A Sigurdsson, Xinlei Chen, and Abhinav Gupta.
\newblock 2016.
\newblock Learning visual storylines with skipping recurrent neural networks.
\newblock In {\em ECCV}.

\bibitem[\protect\citename{Simonyan and Zisserman}2014]{simonyan2014very}
Karen Simonyan and Andrew Zisserman.
\newblock 2014.
\newblock Very deep convolutional networks for large-scale image recognition.
\newblock {\em arXiv preprint arXiv:1409.1556}.

\bibitem[\protect\citename{Spearman}1904]{spearman1904proof}
Charles Spearman.
\newblock 1904.
\newblock The proof and measurement of association between two things.
\newblock {\em The American journal of psychology}.

\bibitem[\protect\citename{Trottier}1998]{trottier1998screenwriter}
David Trottier.
\newblock 1998.
\newblock {\em The screenwriter's bible: A complete guide to writing,
  formatting, and selling your script}.
\newblock Silman-James Press.

\bibitem[\protect\citename{Vendrov \bgroup et al.\egroup
  }2016]{vendrov2015order}
Ivan Vendrov, Ryan Kiros, Sanja Fidler, and Raquel Urtasun.
\newblock 2016.
\newblock Order-embeddings of images and language.
\newblock In {\em ICLR}.

\bibitem[\protect\citename{Wang \bgroup et al.\egroup }2016]{wanglow}
William~Yang Wang, Yashar Mehdad, Dragomir~R Radev, and Amanda Stent.
\newblock 2016.
\newblock A low-rank approximation approach to learning joint embeddings of
  news stories and images for timeline summarization.
\newblock In {\em NAACL}.

\bibitem[\protect\citename{Webber}1988]{webber1988tense}
Bonnie~Lynn Webber.
\newblock 1988.
\newblock Tense as discourse anaphor.
\newblock {\em Computational Linguistics}.

\bibitem[\protect\citename{Weston \bgroup et al.\egroup
  }2015]{weston2015towards}
Jason Weston, Antoine Bordes, Sumit Chopra, and Tomas Mikolov.
\newblock 2015.
\newblock {Towards AI-complete question answering: A set of prerequisite toy
  tasks}.
\newblock {\em arXiv preprint arXiv:1502.05698}.

\bibitem[\protect\citename{Zhu \bgroup et al.\egroup }2015]{zhu2015aligning}
Yukun Zhu, Ryan Kiros, Rich Zemel, Ruslan Salakhutdinov, Raquel Urtasun,
  Antonio Torralba, and Sanja Fidler.
\newblock 2015.
\newblock Aligning books and movies: Towards story-like visual explanations by
  watching movies and reading books.
\newblock In {\em CVPR}.

\end{thebibliography}
	\end{document}